\documentclass[conference]{IEEEtran}
\IEEEoverridecommandlockouts
\usepackage{cite}
\usepackage{amsmath,amssymb,amsfonts}
\usepackage{algorithmic}
\usepackage{graphicx}
\usepackage{textcomp}
\usepackage{xcolor}
\usepackage{hyperref}

\def\BibTeX{{\rm B\kern-.05em{\sc i\kern-.025em b}\kern-.08em
    T\kern-.1667em\lower.7ex\hbox{E}\kern-.125emX}}
\begin{document}

\title{On Importance of Pruning and Distillation for Efficient Low Resource NLP\\
\thanks{L3Cube Labs. Correspondence mail: ravirajoshi@gmail.com}
}

\author{\IEEEauthorblockN{Aishwarya Mirashi}
\IEEEauthorblockA{
\textit{Pune Institute of Computer Technology}\\
\textit{L3Cube Labs}\\
}
\and
\IEEEauthorblockN{Purva Lingayat}
\IEEEauthorblockA{
\textit{Pune Institute of Computer Technology}\\
\textit{L3Cube Labs}\\
}
\and
\IEEEauthorblockN{Srushti Sonavane}
\IEEEauthorblockA{
\textit{Pune Institute of Computer Technology}\\
\textit{L3Cube Labs}\\
}
\and
\IEEEauthorblockN{Tejas Padhiyar}
\IEEEauthorblockA{
\textit{Pune Institute of Computer Technology}\\
\textit{L3Cube Labs}\\
}
\and
\IEEEauthorblockN{Raviraj Joshi}
\IEEEauthorblockA{
\textit{Indian Institute of Technology Madras}\\
\textit{L3Cube Labs}\\
}
\and
\IEEEauthorblockN{Geetanjali Kale}
\IEEEauthorblockA{
\textit{Pune Institute of Computer Technology}\\
Pune, India \\
}
}

\maketitle

\begin{abstract}
The rise of large transformer models has revolutionized Natural Language Processing, leading to significant advances in tasks like text classification. However, this progress demands substantial computational resources, escalating training duration, and expenses with larger model sizes. Efforts have been made to downsize and accelerate English models (e.g., Distilbert, MobileBert). Yet, research in this area is scarce for low-resource languages.

In this study, we explore the case of the low-resource Indic language Marathi. Leveraging the marathi-topic-all-doc-v2 model as our baseline, we implement optimization techniques to reduce computation time and memory usage. Our focus is on enhancing the efficiency of Marathi transformer models while maintaining top-tier accuracy and reducing computational demands. Using the MahaNews document classification dataset and the marathi-topic-all-doc-v2 model from L3Cube, we apply Block Movement Pruning, Knowledge Distillation, and Mixed Precision methods individually and in combination to boost efficiency. We demonstrate the importance of strategic pruning levels in achieving desired efficiency gains. Furthermore, we analyze the balance between efficiency improvements and environmental impact, highlighting how optimized model architectures can contribute to a more sustainable computational ecosystem. Implementing these techniques on a single GPU system, we determine that the optimal configuration is 25\% pruning + knowledge distillation. This approach yielded a 2.56x speedup in computation time while maintaining baseline accuracy levels.

\end{abstract}


\section{Introduction}
A lot of work has already been done in the English language to develop smaller models from large transformer models. One such example is Distillbert \cite{sanh2019distilbert}, a distilled version of BERT \cite{devlin2018bert}, which is 60\% faster and 40\% smaller than the base model while retaining 97\% of its understanding and knowledge. 
However, similar work in Indic languages is still in its nascent stages. Indian regional languages are widely spoken all over India, but small and efficient language models in these languages are not yet available. Hence in this work, we specifically focus on Marathi language and Marathi language models, using marathi-topic-all-doc-v2 {\footnote{https://huggingface.co/l3cube-pune/marathi-topic-all-doc-v2}} as our base model. We try to create a smaller and faster version of this model while trying to retain as much of its language understanding capabilities as possible. 

\par
We implement optimization techniques like Block Movement Pruning, Knowledge Distillation, and Mixed Precision. First, we investigate the efficacy of Block Movement Pruning \cite{lagunas2021block} with different levels of Pruning, to significantly reduce the size of the model. Second, we apply Knowledge Distillation \cite{buciluǎ2006model}, \cite{hinton2015distilling} using the base model as the teacher model and the pruned model as the student model, to further enhance the efficiency of the pruned model while preserving the baseline accuracy. We apply Block Movement Pruning separately and then again in combination with Knowledge Distillation. We perform 6 experiments in total which include:  25\% pruning, 50\% pruning, 75\% pruning, 25\% pruning + knowledge distillation, 50\% pruning + knowledge distillation, and  75\% pruning + knowledge distillation. 

We focus on 3 perspectives while maintaining the baseline accuracy:
\begin{itemize}
    \item Reduction in training time
    \item Reducing carbon dioxide emissions
    \item Computational resources required
\end{itemize}

\subsection{\textbf{Reduction in training time}}
Training time is a critical aspect in the development of transformer models. The efficiency and speed at which a model can be trained directly impact its feasibility for NLP applications.

\subsection{\textbf{Reducing carbon dioxide emissions}}
We calculate the energy efficiency and carbon emissions of training and fine-tuning each model by using a lightweight and easy-to-use Python pip package called CodeCarbon{\footnote{https://codecarbon.io/}. CodeCarbon calculates a weighted average of the emissions from the energy sources that make up our local grid (or the grid used by the cloud provider). When available, CodeCarbon uses the global carbon intensity of electricity per cloud provider or country. The emissions returned by CodeCarbon are only an estimate.

\subsection{\textbf{Computational resources required}}
Training and developing large Transformer Models incur significant financial expenses attributed to hardware, electricity, and cloud computing time. Our experiments show that 25\% pruning combined with knowledge distillation, has 223 million parameters with minimal drop in accuracy. On the other hand, 75\% pruning with knowledge distillation retains 195 million parameters, with a 2\% drop in accuracy. This results in fewer computational resources being consumed for running the models.

\section{Related Work}
This segment discusses advancements in three key areas, discussed earlier, influencing the evolution of transformer models in natural language processing i.e. 
reduction in training time, reducing carbon dioxide emission and computational resources required. In pursuit of more effective and sustainable AI development, researchers are delving into techniques such as knowledge distillation,  pruning, model quantization, and mixed precision. These approaches aim to expedite training processes while upholding model accuracy. The Lottery Ticket Hypothesis by \cite{frankle2018lottery} inspired wide research in the field of training sparse neural networks.

Large language models can be resource intensive, but Fastformers \cite{kim2020fastformers} offers a solution. It uses a combination of three techniques to significantly speed things up, like giving the model shortcuts to process information. In tests, Fastformers ran way faster than standard models - up to 234 times faster on CPUs and 12.4 times faster on GPUs. However, they found that speed comes at the cost of some accuracy. For instance, removing unnecessary parts with pruning made the model 3 times faster on CPU, but resulted in a small drop in accuracy. Fastformers address this by carefully removing parts of the model using structured pruning while keeping accuracy high. Then, knowledge distillation is used to further improve the accuracy of the pruned model. Finally, the model uses a more efficient way of representing numbers of lower precision arithmetic to squeeze even more speed out of it, while maintaining acceptable accuracy.

\cite{lakim2022holistic} studied the environmental impact of creating four large language models called Noor models. They looked at everything from collecting and storing data to training the models and how they'll be used in the future. In total, creating these models produced an estimated 36.5 tons of carbon dioxide. Most of the emissions (65\%) came from training the models. The rest came from things like travel for training (18\%), data storage (12\%), and smaller research experiments (4\%).

\cite{schwartz2020green} advocates for disclosing the financial costs associated with developing, training, and running models, termed the "price tag." They argue that such transparency is essential for establishing benchmarks in the quest for more efficient methods. The authors propose that research papers should include analyses plotting accuracy against computational cost and training set size to provide a reference for future research on data efficiency. In their view, three main factors contribute to the computational and environmental costs of producing results: The cost per example (E) during model execution, whether in training or inference. The size of the training dataset (D), which influences how frequently the model is executed during training. The number of hyper-parameter experiments (H) conducted during model development, affects how often the model is trained. They emphasize that the overall cost of achieving results in machine learning increases linearly with these factors. \cite{strubell2019energy} performed a comparative analysis of the energy required to train a variety of popular off-the-shelf NLP models. They concluded that authors should report training time and sensitivity to hyper-parameters, academic researchers need equitable access to computation resources and researchers should prioritize computationally efficient hardware and algorithms.

\cite{sanh2020movement} introduces movement pruning, a novel technique for making large language models (LLMs) smaller and more efficient during fine-tuning. The paper compares movement pruning with other pruning techniques on various fine-tuning tasks. Movement pruning produces 77\% savings in parameter storage for a 1\% drop in accuracy on SQuAD v1.1. The results show that movement pruning consistently achieves better sparsity-accuracy trade-offs, meaning it can significantly reduce model size while maintaining high accuracy. Notably, in high-sparsity regimes where a large portion of the model is pruned, movement pruning when combined with distillation, can reduce the model size to just 3\% of its original size with minimal accuracy loss.

Distillation methods have been used to produce faster models. DistilBERT \cite{sanh2019distilbert} leverages knowledge distillation during the pre-training phase and distills BERT into a 40\% smaller model. TinyBERT \cite{jiao2019tinybert} distills a fine-tuned model by using data augmentation. MobileBERT \cite{sun2020mobilebert} is the result of a large architecture search. These approaches utilize targeted distillation to produce smaller models with a dense structure that is fast on standard hardware. 

\cite{lagunas2021block} proposes an approach that encourages pruning that can be optimized on dense hardware. They extend movement pruning to work on blocks of local parameters. Their experiments showed that despite utilizing sub-row square blocks during training, the approach learns to eliminate full components of the model, effectively dropping a large number of attention heads. This effect allows the model to achieve speedups even beyond standard structured pruning of feed-forward layers. They obtained a pruned model that is a 2.4x faster, 74\% smaller BERT on SQuAD v1, with a 1\% drop on F1

\cite{micikevicius2017mixed} discusses training deep neural networks with mixed precision. The authors propose using half-precision floating point numbers to reduce memory consumption during training. However, half-precision numbers have a limited numerical range, which can cause accuracy issues. To address this challenge, the paper introduces two techniques:

\begin{enumerate}
    \item Maintaining a Single-Precision Copy of Weights: 

    \item Scaling the Loss: 
\end{enumerate}

They demonstrated that training deep neural networks using half-precision floating point numbers can reduce the memory consumption of deep learning models by nearly 2x. This technique works for large-scale models with more than 100 million parameters trained on large datasets.

\cite{jain2020indic} investigates the use of Transformer-based language models for Indian languages. The authors analyze these models by performing experiments on Hindi, Bengali, and Telugu. They compare two approaches: fine-tuning pre-trained models and training a model from scratch. Their findings suggest that the best approach depends on the specific task, and not necessarily the size of the dataset available. The paper achieves state-of-the-art performance on text classification tasks for Hindi and Bengali.

\cite{kim2023adapt} The paper proposes a strategy (Adapt and Prune) to improve the performance of large speech models for Korean, Malayalam, Japanese, Swahili and Chinese. This strategy involves Pruning, where the model identifies and removes neurons that are not useful for understanding the specific low-resource language and Adapting, where a small, additional module is added to the reduced model. This module is trained specifically on the low-resource language data, allowing the model to adapt to the unique characteristics of that language. The authors propose a new adaptation method called PEPSI (Pruning with Enhanced fine-tuning for Speech Integration). Experiments showed that PEPSI outperforms the commonly used LoRA (Low-Rank Adaptation) technique and matches the performance of Full-Fine Tuning (FFT) while significantly reducing the number of parameters (up to 50\% for some languages).

\cite{ahia2021low} explores a challenge in machine translation called the "low-resource double bind." This situation occurs when we are working with languages that have both limited data (low-resource) and limitations on computing power. The authors introduce a new technique called magnitude pruning to address this problem. Magnitude pruning focuses on removing connections with very small values.
They investigate how magnitude pruning affects machine translation in low-resource settings. They found that pruning can:
\begin{itemize}
    \item Maintain performance on frequently occurring sentences.
    \item Negatively impact the translation of less frequent sentences.
    \item Improve the model's ability to handle situations where the translated text is very different from what the model was trained on.
\end{itemize}
Overall, the paper suggests that pruning can be a valuable tool for machine translation tasks with limited data and computing resources. It helps prevent the model from memorizing unimportant details and improves its generalizability.

\begin{figure*}[hbt!]
    \centering
    \includegraphics[width=\linewidth]{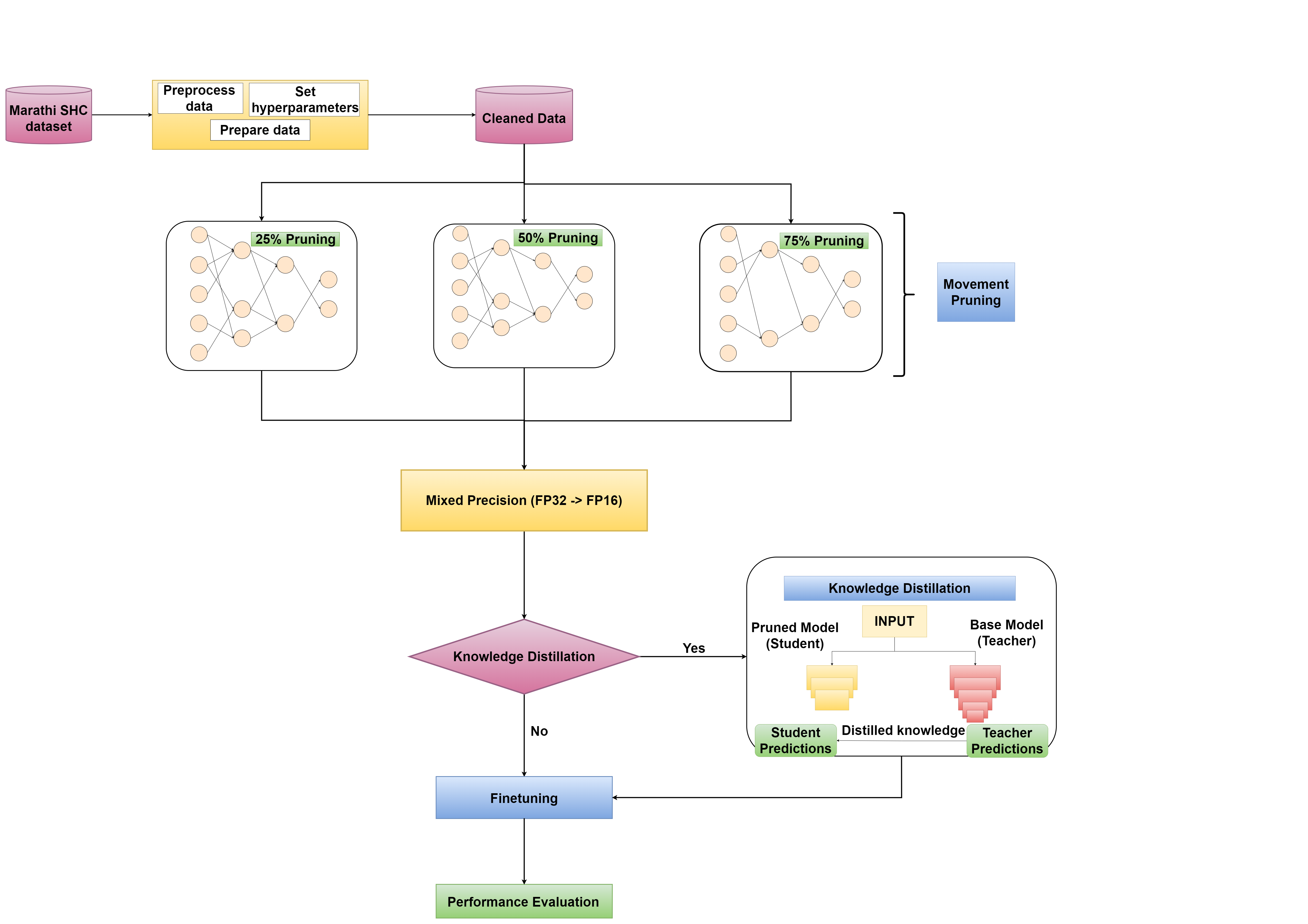}
    \caption{Proposed model flow}
    \label{fig:enter-label}
\end{figure*}

\section{Optimization Techniques}
In this section, we explore the different optimization techniques to enhance the training efficiency of large transformed models. We mainly study 3 techniques- Block Movement Pruning, Knowledge Distillation, and Mixed Precision. 

\subsection{\textbf{Block Movement Pruning}}
    Movement pruning \cite{sanh2020movement} is an effective learning-based unstructured sparsification algorithm, especially for Transformer-based models in transfer learning setup. It is a technique used to train neural network models with sparse connections dynamically during the training process. Movement pruning selectively prunes and restores connections during each training iteration based on certain criteria. High levels of sparsity can be reached with a minimal accuracy loss. It is a very efficient method to prune networks in an unstructured manner. 
    \par
    \cite{lagunas2021block} further studied movement pruning \cite{sanh2020movement} and found that without careful engineering and size selection, these models are much larger than pruned ones. They solve this issue through block pruning. Block movement pruning \cite{lagunas2021block} is a weight pruning method that extends the original movement pruning method by exploring structured and semi-structured variants. It extends the movement pruning algorithm to sparsify by block grain size, enabling structured sparsity which can achieve device-agnostic inference acceleration.

    \cite{zhu2017prune} dives into the effectiveness of pruning. Their key finding states that large but pruned models (large-sparse) can outperform smaller, dense models (small-dense) with the same memory footprint. This challenges the traditional notion that smaller models inherently perform better. The paper reports reductions in the number of parameters by up to 10 times with minimal accuracy loss through pruning.

    \cite{sun2023simple} introduces a new pruning method called Wanda (Pruning by \textbf{W}eights \textbf{and} \textbf{a}ctivations), specifically designed for compressing LLMs efficiently. Wanda considers both weights and activations. The authors show that Wanda achieves good results without retraining and is effective across different sparsity levels. Wanda significantly outperforms the established baseline of magnitude pruning and performs competitively against recent method involving intensive weight update.

\subsection{\textbf{Knowledge Distillation}}
    Knowledge distillation \cite{buciluǎ2006model}, \cite{hinton2015distilling} is a compression technique, where a larger model (teacher) is used to train a smaller model (student). The student model tries to replicate the behavior of the teacher model. The goal is to distill the knowledge of the teacher into the student, allowing the student to achieve performance similar to the teacher while being more computationally efficient. 

    \cite{cui2017knowledge} explores using knowledge distillation to improve performance in machine translation for low-resource languages. Instead of a single teacher, the authors propose using an ensemble of multilingual models as teachers. This ensemble approach potentially captures a wider range of knowledge beneficial for the student model. Their research finds that the student model trained with knowledge distillation can even outperform the teacher models and the original ensemble.  
    
    In the last 3 experiments, we apply Block Movement Pruning followed by Knowledge Distillation.
    In this case, the teacher model is our base model, marathi-topic-all-doc-v2 and the student model is the pruned model. 
    
\subsection{\textbf{Mixed Precision}}
    Mixed precision training \cite{micikevicius2017mixed} is a technique that aims to optimize the computational efficiency of training models by utilizing lower-precision numerical formats for certain variables. Traditionally, most models use 32-bit floating point precision (fp32 or float32) to represent and process variables. By reducing the precision of certain variables to lower numerical formats like 16-bit floating point (fp16 or float16), we can speed up the computations.

\section{Dataset and Model}
The Marathi dataset we used is \footnote{https://github.com/l3cube-pune/MarathiNLP/tree/main/L3Cube-MahaNews}. The dataset was curated by scraping Marathi news websites. The dataset statistics are as follows: Training set: 22014 records, Validation set: 2750 records, and Test set: 2751 records. The base model used is marathi-topic-all-doc-v2. \cite{mirashi2024l3cube}. The model has been trained and finetuned on the L3Cube MahaNews dataset. 

\begin{figure}[hbt!]
\centering
    \includegraphics[width=0.9\linewidth]{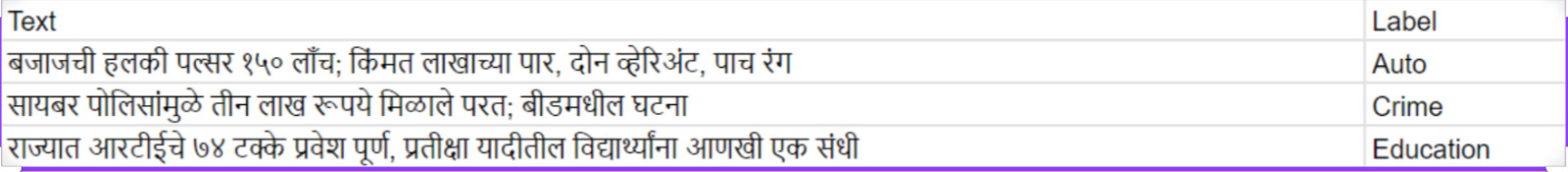}
    \caption{Dataset Overview}
    \label{fig:enter-label}
\end{figure}

\section{Experimental setup}
In this section, we dive into our implementation of the 3 proposed techniques. We implemented the techniques in 2 parts. In the first part, we conduct 3 experiments, performing only Block Movement Pruning by pruning 25\%, 50\%, and 75\% of the base model. Later, in the second part, we conduct 3 more experiments, wherein we perform knowledge distillation on the pruned model, using the base model as our teacher model and the pruned model as our student model. The pruned model learns from the base model and tries to achieve performance similar to the marathi-topic-all-doc-v2 model.

\begin{table*}[hbt!]
    \centering
    \caption{Results of 6 optimization experiments on GPU}
    \resizebox{\textwidth}{!}{%
    \begin{tabular}{|p{3.5cm}|p{1.5cm}|p{1.8cm}|p{2cm}|p{1.3cm}|p{2.2cm}|}
    \hline
        \textbf{Optimization added} & \textbf{Number of parameters (in million)} & \textbf{\% Reduction in no of parameters} & \textbf{\% decrease in inference time/ Time Speed up} & \textbf{Accuracy} & \textbf{Estimated Carbon emissions (kg CO2 eq) } \\
        
        \hline
        {Baseline} & 238 & - & - &\textbf{92.43}& 0.006135 \\
        \hline
        {25\% pruning} & 223 & 5.96 & 30.11 (1.43x) & \textbf{92.29} & 0.002735 \\
        \hline
        {50\% pruning} & 209 & 12 & 40.21(1.67x) & 91.81 & 0.002663 \\
        \hline
        {75\% pruning} & 195 & 18 & 55.6(2.25x) & 90.11 & 0.001335 \\
        \hline
        {25\% pruning + knowledge distillation} & 223 & 5.96 & \textbf{59.06(2.56x)} & \textbf{92.18} & 0.003464 \\
        \hline
        {50\% pruning + knowledge distillation} & 209 & 12 & 56.98(2.32x) & 91.49 & 0.003301 \\
        \hline
        {75\% pruning + knowledge distillation} & 195 & 18 & \textbf{55.6(2.25x)} & 90.55 & 0.003723 \\
        \hline        
    \end{tabular}
    }
\end{table*}


We use the methodologies proposed by \cite{lagunas2021block} and \cite{sanh2020movement}, to implement block movement pruning. We initially started by setting the pruning values to 25\%, 50\%, and 75\% pruning, meaning that the final value of the masking threshold was set to 0.75, 0.5 and 0.25 respectively. This allowed us to evaluate the pruned model at different levels of pruning. The marathi-topic-all-doc-v2 model contains 238 million parameters. In our experiments, we were able to retain 223M, 209M, and 195M parameters with 25\%, 50\%, and 75\% pruning respectively. Our experiments demonstrated a 1.43x speedup with 25\% pruning, a 1.67x speedup with 50\% pruning, and a 2.25x speedup with 75\% pruning from the computational time required for the base model.

While pruning significantly reduces the model size, knowledge distillation is used to speed up the inference. In the last 3 experiments, we applied knowledge distillation to the pruned model that we obtained in the first part. The base model, marathi-topic-all-doc-v2 is used as the teacher model and the pruned model is used as the student model. 
\cite{hinton2015distilling} demonstrated that knowledge distillation can successfully transfer knowledge from a large, complex "teacher" model to a smaller, more efficient "student" model. It also showed that Knowledge Distillation performs best when the teacher and student are of the same model type. So in our work, we use marathi-topic-all-doc-v2, which is a BERT-based model, and the pruned model, which is also a BERT-based model. We observed that for 25\% pruning followed by Knowledge Distillation, the speed up in computational time goes from 1.43x to 2.56x and for 50\% pruning with Knowledge Distillation, the speed up in computational time goes from 1.67x to 2.32x. For 75\% pruning with Knowledge Distillation, there was no change in the speed up. 

In our work, the pruned model containing a significantly lesser number of parameters than the base model, learned to replicate the behavior of the base model and achieve performance similar to marathi-topic-all-doc-v2 on the MahaNews dataset. 
\cite{micikevicius2017mixed} observed that using half-precision floating point numbers, deep neural networks can be trained, without losing model accuracy. It proposed converting single-precision outputs into half-precision before storing them in memory. We perform mixed precision training while training the pruned model. We convert single-precision floating-point variables (fp32) into half-precision floating-point variables (fp16).


\begin{figure}[hbt!]
    \centering
    \includegraphics[width=\linewidth, height=5.5cm]{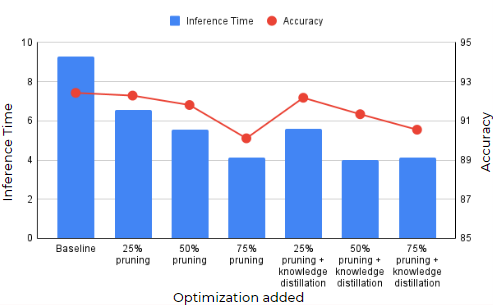}
    \caption{Accuracy vs. Time with various optimizations on GPU}
    \label{fig:enter-label}
\end{figure}

\section{Results}

Table 1 represents the 6 experiments performed along with the number of parameters in each model, its accuracy, and approximate carbon emissions released (in kg CO2 eq) from running the model. The final number of parameters in each model depends on the level of pruning performed.

The base model has 238 million parameters. On 25\% pruning, the number of parameters gets reduced by 5.96\%, to 223 million parameters. Similarly, on 50\% pruning, the number of parameters gets reduced by 12\% to 209 million parameters. On pruning 75\% of the weights, the number of parameters gets reduced by 18\% to 195 million parameters.

\section{Conclusion}
In this paper, we have enhanced the efficiency as well as reduced the size of the marathi-topic-all-doc-v2 model by using optimization techniques like Block Movement Pruning, Knowledge Distillation, and Mixed Precision Training, using 6 different experiments. The resulting models achieve comparable accuracy on the MahaNews dataset while exhibiting significant reductions in model size, inference time, and carbon footprint. These findings underscore the importance of environmentally responsible NLP model development. By optimizing architectures and minimizing computational demands, we not only achieve faster processing but also contribute to a greener AI ecosystem. Our research offers valuable insights for researchers and practitioners, highlighting the efficacy of efficiency enhancement strategies in NLP tasks. We advocate for continued exploration of innovative approaches that prioritize both performance and sustainability, paving the way for a more impactful and environmentally conscious future of NLP research and applications. 

\section*{Acknowledgments}

This work was done under the L3Cube Labs, Pune mentorship program. We would like to express our gratitude towards our mentors at L3Cube for their continuous support and encouragement.

\bibliography{main}
\bibliographystyle{ieeetr}

\end{document}